\documentclass[sigconf, natbib=false]{acmart}

\usepackage{cite}
\usepackage{amsmath,amsfonts}

\usepackage{algorithmic}
\usepackage{booktabs} 
\usepackage{siunitx}
\usepackage{tabularx}
\usepackage{graphicx}
\usepackage{subcaption}
\usepackage{textcomp}
\usepackage{xcolor}

\AtBeginDocument{%
  \providecommand\BibTeX{{%
    \normalfont B\kern-0.5em{\scshape i\kern-0.25em b}\kern-0.8em\TeX}}}

\acmConference[Conference acronym KDD]{ACM Knowledge Discovery and Data Mining}{25th - 29th August,
  2024}{Barcelona, Spain}

\acmISBN{978-1-4503-XXXX-X/18/06}

\begin{document}

\title{Empowering Data Mesh with Federated Learning}

\author{Haoyuan Li}
\affiliation{%
  \institution{Department of Industrial Design,\\ Eindhoven University of Technology}
  \city{Eindhoven}
  \country{Netherlands}}
\email{h.y.li@tue.nl}

\author{Salman Toor}
\affiliation{%
  \institution{Department of Information Technology,\\ Uppsala University \\ Scaleout Systems}
  \city{Uppsala}
  \country{Sweden}}
\email{Salman.Toor@it.uu.se }
\email{Salman@scaleoutsystems.com}

\begin{abstract}
The evolution of data architecture has seen the rise of data lakes, aiming to solve the bottlenecks of data management and promote intelligent decision-making. However, this centralized architecture is limited by the proliferation of data sources and the growing demand for timely analysis and processing. A new data paradigm, Data Mesh, is proposed to overcome these challenges. Data Mesh treats domains as a first-class concern by distributing the data ownership from the central team to each data domain, while keeping the federated governance to monitor domains and their data products. Many multi-million dollar organizations like Paypal, Netflix, and Zalando have already transformed their data analysis pipelines based on this new architecture.  
In this decentralized architecture where data is locally preserved by each domain team, traditional centralized machine learning is incapable of conducting effective analysis across multiple domains, especially for security-sensitive organizations. To this end, we introduce a pioneering approach that incorporates Federated Learning into Data Mesh. 
To the best of our knowledge, this is the first open-source applied work that represents a critical advancement toward the integration of federated learning methods into the Data Mesh paradigm, underscoring the promising prospects for privacy-preserving and decentralized data analysis strategies within Data Mesh architecture.
\end{abstract}


\keywords{Data Mesh, Federated Learning, Domain-Intelligence, Privacy, Data-driven analysis}

\maketitle

\section{Introduction}
The concept of Big Data has experienced significant evolution since its introduction, with continuous advancements over the past few decades. Initially characterized by the $3Vs$ model (Volume, Variety, Velocity), as proposed by Doug Laney \cite{beyer2012importance}, this model has expanded to encompass Veracity and Value, culminating in the $5Vs$ model to describe the main challenges when handling big data. In today's data-driven landscape, the rapid growth of data volume and complexity in the modern world poses significant challenges for data-driven organizations that are striving to explore the potential value of their data asset\cite{datameshblog2019}. \par

In response to these challenges, concepts such as Data Warehouses and Data Lakes have emerged, providing structured and unstructured repositories, respectively, that can store, manage, and analyze vast amounts of data \cite{khine2018data}. In this monolithic data paradigm, decision-making relies on a centralized data team to process and manage the analytical data. However, many organizations are suffering from this centralized pattern when scaling to accommodate large volumes of data \cite{datameshblog2019, bode2023data}. The central data team faces considerable challenges as it bears the responsibilities associated with managing diverse data types and meeting the escalating demands of geographically dispersed business units, a predicament further intensified by the global expansion of modern enterprises. Consequently, organizations often find themselves dedicating substantial time and resources to resolving data silos issues, hindering timely and effective data-driven decision-making.\par


To overcome this bottleneck, a paradigm shift toward decentralized data management has been proposed, known as the Data Mesh architecture \cite{datameshblog2019, datameshblog2020}. This novel architecture advocates for the distribution of responsibilities and ownership of data, enabling organizations to handle large-scale data management in a more efficient and effective manner. Data Mesh embraces the nature of operational data and analytical data, with a focus on domain-oriented, distributed analytical data and federated governance. By integrating product thinking into data management, the Data Mesh promotes the concept of distributed data products, each owned by a specific domain team within an organization, to be consumed by other domains or end-users. As a result, data management becomes a fundamental aspect of the domain teams' responsibilities, encouraging a culture of ownership and accountability to ensure the quality of data products.\par

In the context of decentralized data architecture, data is owned by each domain team, with no requirement for sharing with other domain teams or aggregation into global data lakes. These domain-specific model teams are responsible for the production of machine learning models to serve as interfaces for consumers.  Simultaneously, when organization-wide decisions need to be made, all domain teams are required to collaborate on the project and generate a global model to assist the federated governance team. To achieve this objective, we incorporate federated learning in our study to train the machine learning models in a decentralized manner.\par

The topology of federated learning models aligns well with the Data Mesh structure, as it inherently supports the decentralization and domain-specific governance principles of the Data Mesh. Federated learning enables each domain team to train machine learning models without accessing raw data in other domains, respecting the principle of data locality and ownership, which is a cornerstone of the Data Mesh architecture. 
Due to the domain-oriented architecture, federated learning facilitates the creation of more robust and diverse models by enabling learning from a variety of domain-specific datasets. In the context of Data Mesh, where each domain team owns different types of data, the models trained through federated learning can potentially benefit from this diversity, leading to more generalized and accurate predictions. The proposed solution is based on an open-source applied work toward the integration of federated learning methods into the Data Mesh paradigm. \par

The following list is the main contribution of this work:
\begin{enumerate}
    \item Identifying the main characteristics of machine learning models when conceptualized as data products within a distributed data architecture. 
    
    
    \item Constructing the domain-specific architecture of the split learning model under scenarios of both shared and preserved labels.
    
    \item Propose two common use cases that demonstrate the advantages of domain-oriented data segregation for business applications. 
\end{enumerate}

\section{Related Work}

The Data Mesh concept was initially proposed by Zhamak Dehghani in a foundational blog post, where it was positioned as a response to the limitations of prevailing monolithic data architectures. The blog articulated the need for a paradigm shift towards decentralized data structures, laying out the key motivations behind the Data Mesh concept and summarizing the fundamental attributes of the logical components within the data mesh \cite{datameshblog2019}. A subsequent post further elaborated on the Data Mesh, outlining four primary principles: domain ownership, data as a product, a self-serve data platform, and federated computational governance \cite{datameshblog2020}. These principles form the backbone of a comprehensive high-level Data Mesh architecture, offering a standard logical model for further investigation. \par

Building upon this architectural foundation, numerous businesses have reported practical implementations of the Data Mesh, adapting it to align with their specific business strategies. For instance, Zalando \cite{zalandoURL}, an e-commerce enterprise dealing with a myriad of data sources, has engineered a tailored Data Mesh framework to alleviate the bottlenecks of a central team and ensure data quality. This framework incorporates the "Bring Your Own Bucket" (BYOB) mechanism to facilitate decentralized data storage. 
Users are able to integrate their data buckets with the centralized data lake, allowing them to leverage processing platforms or techniques offered by the governance team \cite{zalando, MACHADO2022263}. 
Similarly, Netflix \cite{netflixURL} has designed its own Data Mesh platform to enhance data movements within Netflix Studio. A primary focus of this platform is to establish a self-service environment where users can develop their data pipelines and adhere to standardized process policies, thereby minimizing redundant effort in data pipeline development. 
The platform is built upon the context of the ETL (Extract-Transform-Load) process in a self-service manner. Additionally, it applies schemas to all data pipelines to guarantee data event quality and simplify data discovery \cite{netflixblog1, netflixblog2, NetflixVideo}. \par

Other data-driven companies, like PayPal \cite{paypalURL} and Intuit \cite{intuitURL}, are also working on the implementation of their strategies for Data Mesh, taking their initial steps on the Data Mesh journey. As a pioneer in self-service analytics, PayPal decomposes the architecture of Data Mesh into a set of deployable elements called data quantum maintained by the specific domain data team \cite{paypaldatamesh}. 
Intuit also provided its own visions for Data Mesh, formulating the future direction of its data-driven systems. They define a set of strategies from three perspectives that can improve the process of data discovery and organization \cite{intuitdatamesh}. 
\par

The Data Mesh paradigm is gaining momentum as a promising solution for businesses struggling with the constraints of conventional data management approaches. An expanding body of research is devoted to delivering practical guidelines for implementing a Data Mesh from an enterprise perspective.  In their work, Butte et al. \cite{butte2022enterprise} concentrate on constructing domain components, addressing interoperability amongst these entities. They further furnish an abstract cloud service-based architectural blueprint for implementing the Data Mesh. Bode et al. \cite{bode2023data} provide empirical insights into the Data Mesh, derived from $15$ semi-structured interviews conducted with industry experts. 
Machado et al. propose the domain model to represent the basic components that reside on each independent domain \cite{machado2021data}. 
The evolution of the modern data paradigm is identified in \cite{MACHADO2022263}, together with the features and deployment strategy for Data Mesh.

\section{Core Components}

\subsection{Data Mesh}

Data Mesh emphasizes the decentralization of data ownership, domain-oriented data products, self-serve data infrastructure, and federated computational governance. This empowers domain teams to take ownership of their data while fostering collaboration and data sharing across domains. Notably, the principle of the self-serve data infrastructure, while essential in understanding the entirety of Data Mesh, falls outside of this research's scope due to its distinct focus on infrastructural considerations that do not directly interact with our exploration of machine learning methods. In the following sections, we present the overview of Data Mesh principles in our study, highlighting the essential aspects for training machine learning models under this decentralized architecture.

\subsubsection{Domain-Oriented Data Ownership}

The core feature of Data Mesh is decentralized data management. Drawing inspiration from Domain-Driven Design (DDD) principles, Data Mesh distributes data ownership across various domains\cite{datameshblog2019}. The centralized analytical data is distributed to domain-oriented teams responsible for managing and processing their domain data. These domain teams can create and maintain data products for end-users or other domains. Furthermore, domain teams can collaborate on global activities under the guidance of a global team, enabling efficient and flexible data management across the organization.

\subsubsection{Data as a Product}

The underlying thought of this principle comes from combining product thinking with analytical data. The domain team, as the data owner, should assume responsibility for generating and sustaining data products for consumers. Structurally, a well-defined data product comprises four components: code, data and metadata, infrastructure, and interfaces\cite{datameshblog2020}. 
Depending on the source and purpose, data products can be categorized into two types: atomic data products and composite data products. Atomic data products originate from source data and cater to end-users or downstream data products. Composite data products are formed by ingesting data from other data products, encompassing both basic and composite data products from upstream sources. These data products are designed to be consumed by end-users rather than other domain teams for specific use cases. 

Although customized data products are produced for various purposes, they all share similar characteristics to ensure their quality. In our study, each domain team cooperates to train the split neural network, and each domain produces a partial model to serve the global server. We use eight attributes identified by \cite{goedegebuure2023data} to present an overview of a high-quality product generated by federated learning (model as a product).

\subsubsection{Federated Computational Governance}

To balance the concentration between centralization and decentralization,  Data Mesh creates a global-level entity inside its architecture to govern each domain. This federated governance model aims to facilitate efficient collaboration and coordination among domain teams while maintaining a high level of autonomy. It defines the global standardization rules to ensure the interoperability of each domain. By setting a set of governance policies, the global team can monitor the data products produced by domain teams. 

Federated governance activities in Data Mesh can be categorized into two types: global governance and local governance. Global governance occurs at a higher level within the data mesh and guides domain teams in fulfilling their responsibilities for managing domain-specific data. Meanwhile, local governance takes place closer to the data domain and focuses on maintaining the quality of data products produced by domain teams. In our study, we identified five governance activities \cite{goedegebuure2023data} in data mesh when training federated learning models.\par

\subsection{Federated Learning}

Federated Learning is a machine learning approach that encourages model training across a broad network of independent, decentralized nodes. These nodes, in the context of the data mesh, correspond to the variety of domains where data naturally resides. This methodology aligns closely with the fundamental tenets of the data mesh, offering a host of benefits and making it a suitable choice for machine learning applications within this distributed structure.\par

A key advantage of Federated Learning is its harmony with the philosophy of decentralized data ownership specific to domains, which is a foundational aspect of the data mesh model. In contrast to the issues associated with data copying in centralized learning, Federated Learning allows data to stay in its original domain throughout the learning phase. This practice effectively minimizes the requirement for data duplication and transfer, addressing the associated inefficiencies and potential risks related to data integrity. Furthermore, FL enhances the role of domain owners in the machine learning process. By training models within their respective domains, domain owners can exercise control and provide input into the learning process. This not only potentially improves the quality and relevance of the models but also aligns with the data-as-product principle, ensuring data is managed and curated within its domain context. Moreover, FL addresses privacy and security concerns that are often inherent in centralized learning. By maintaining data within its domain during model training, sensitive data does not need to be exposed to a central authority, reducing the risk of data breaches and privacy violations.\par

Building on these advantages, we will now delve into three distinct types of Federated Learning: Horizontal FL, Vertical FL, and Split Learning. Each of these Federated Learning methods presents unique characteristics that could potentially be beneficial in distributed data architecture. In our study, we concentrate on Split Learning, highlighting the features of Split Learning that align with data mesh and the procedural strategies for its effective implementation within such a framework.

\subsubsection{Horizontal Federated Learning}

\begin{figure*}[htbp]
    \centering
    \includegraphics[width=0.85\textwidth]{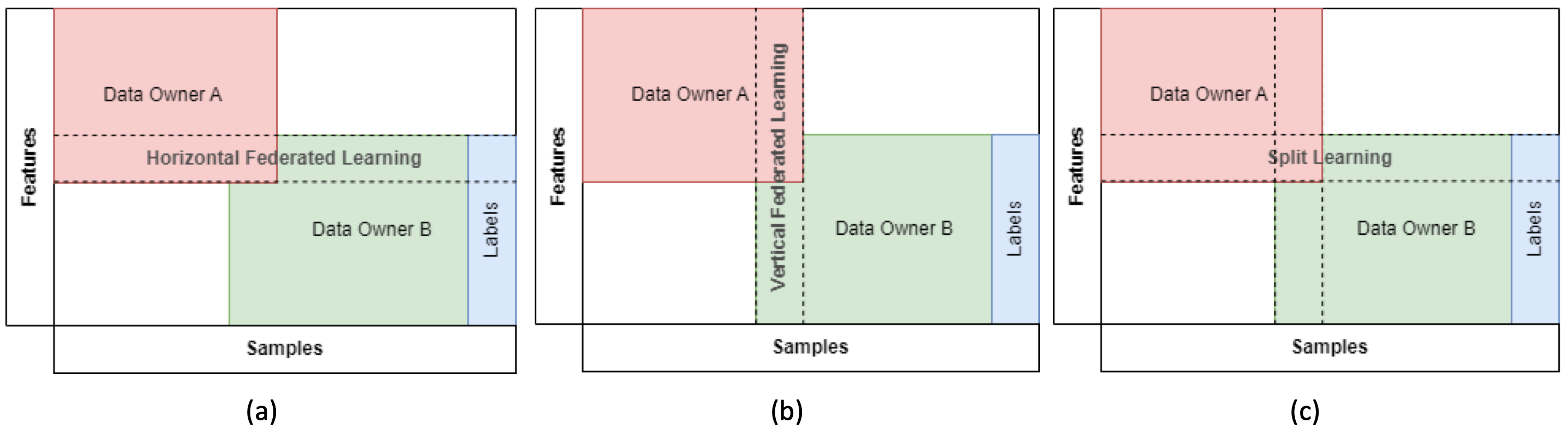}
    \caption{The difference between horizontal federated learning(a), vertical federated learning(b) and split learning(c)}
    \label{fig:FL-usecase}
\end{figure*}

Horizontal Federated Learning (HFL) was first introduced by Google, aiming to train the machine learning models on decentralized data across multiple devices, reducing the need for data transfer and thus enhancing privacy and efficiency.
HFL also known as collaborative learning, is utilized in the scenario where each client or node in the federated learning setup has data from many users, but the feature space is the same or similar across all clients \cite{yang2019federated}, as described in Figure \ref{fig:FL-usecase}-(a). 
\par

In a data mesh architecture, data ownership is distributed across disparate domains, each possessing distinct types and features of data. This decentralized data distribution is contrasted by the assumptions of HFL, which posits that all nodes or clients share a common feature space, differing only in samples or users. 
While HFL maintains its merits and applicability in certain contexts, it may not align optimally with the principles and practicalities of a data mesh environment. Specifically, the data within Data Mesh is distributed to each independent domain and shares similar samples but contains different feature spaces.

\subsubsection{Vertical Federated Learning}

Vertical Federated Learning (VFL) provides a framework that allows different entities to engage in the collaborative training of machine learning models while maintaining robust data privacy and security safeguards \cite{yang2019federated}. 
As is shown in Figure \ref{fig:FL-usecase}-(b), this approach is specially tailored for scenarios wherein participating entities possess disparate feature spaces for identical samples. Such a structure is particularly suitable when direct data sharing is unfeasible due to legal restrictions or privacy considerations.


VFL brings many advantages in terms of data privacy and decentralized learning, but it does encounter certain challenges when applied in a Data Mesh environment. 
VFL necessitates precise data alignment across domains, requiring all domains to maintain identical entities, differing only in the features they hold. However, in a data mesh setting, this level of synchronization might not always be feasible or efficient, thus posing a challenge for VFL's implementation. 

\subsubsection{Split Learning}

Split Learning (SL) is another novel approach that allows training deep neural networks on data from multiple parties in a distributed manner. SL is introduced to resolve security concerns when training deep neural networks for data-sensitive applications \cite{gupta2018distributed}. The key idea behind SL is to perform model training across multiple nodes while minimizing the data communication overhead and preserving data privacy.\par


Consider a neural network composed of $L$ layers.  In SL, we divide this network at the $k$-th layer, this layer is also called the cut layer. The client controls the layers from $1$ to $k$, whereas the server manages the layers from $(k+1)$ to $L$. We symbolize the output from the $k$-th layer as $h_k(x; W_k)$, where $x$ represents the input data, and $W_k$ stands for the model's parameters up to the $k$-th layer. \par

During forward propagation, the client conveys $h_k(x; W_k)$ to the server. The server then employs its section of the model, denoted as $h_{k+1:L}(h_k(x; W_k); W_{k+1:L})$, to calculate the output. Then the loss will be calculated based on the labels on clients, or the server depending on where the labels data is preserved. When backpropagation takes place, the server calculates the gradients concerning its parameters and the input it received, specifically, $\nabla_{W_{k+1:L}}$ and $\nabla_{h_k(x; W_k)}$. The gradient $\nabla_{h_k(x; W_k)}$ is then transmitted back to the client. The client utilizes this gradient to compute and update its parameters using the gradient descent method. The weight update process on the client's end can be formulated as:
\begin{equation}
W_k := W_k - \alpha \cdot \nabla_{W_k} Loss(h_{k+1:L}(h_k(x; W_k); W_{k+1:L}))
\end{equation}
The server's weight update step is represented as:
\begin{equation}
W_{k+1:L} := W_{k+1:L} - \alpha \cdot \nabla_{W_{k+1:L}} Loss(h_{k+1:L}(h_k(x; W_k); W_{k+1:L}))
\end{equation}

\begin{figure}[htbp]
    \centering
    \includegraphics[width=0.30\textwidth]{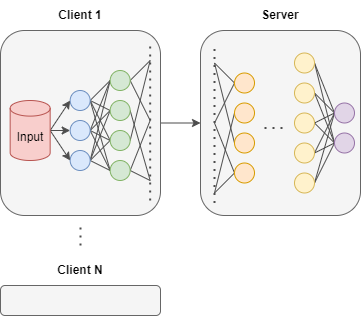}
    \caption{Basic Structure of Split Learning}
    \label{fig:SL}
\end{figure}

In SL, the computation of a neural network model is partitioned into two parts: a client model that processes the initial layers of the network, and a server model that handles the subsequent layers. The raw data is kept within the client side, and only the intermediate representation generated at the cut layer can share with the server for further processing. From a practical perspective, Split Learning is flexible and adaptable, able to support a wide range of network architectures and machine learning tasks. Typically, SL can be tailored for both horizontal and vertical cases based on the distribution of data on the connected clients, as is described in Figure \ref{fig:FL-usecase}-(c). 

Within the architecture of data mesh, Split Learning aligns well with the principle of domain-oriented decentralized data ownership. 
The approach also substantially reduces the amount of data that needs to be transmitted over the network, thereby increasing efficiency and preserving bandwidth. In terms of security, only the intermediate representations, or features extracted from the raw data, are shared for the training process. This reduces the exposure of sensitive raw data, as the shared representations often do not carry explicit sensitive information, or they make it substantially harder to extract \cite{vepakomma2018split}. \par


\section{System Architecture}


In our research, we propose two distinct architectures for two separate scenarios within the geographically distributed data mesh - label sharing and label preserving. These designs strictly adhere to the domain-oriented principle whereby each domain is recognized as an independent data owner. Each domain adheres to a no-peek policy, which limits its access to raw data from other domains. This policy, however, permits the sharing of data products like intermediate model weights or gradients exchanged among domains. For our experimental setup, we employed a concatenation-based aggregation mechanism. While this approach is simple and targeted, it is not immune to the issue of stragglers. Nonetheless, there are alternative strategies, such as element-wise sum and average pooling, which could be explored in varying contexts\cite{ceballos2020splitnn}.

\subsection{Distributed Domain Data with Label Sharing}

In the first scenario, label data is securely retained by the consumer located on the server side. The loss calculation is executed server-side, and subsequently, gradients are back-propagated. These gradients follow the sequence layers, being disseminated to each respective domain model at the point of the aggregation layer. 

\begin{figure}[htbp]
    \centering
    \includegraphics[width=0.40\textwidth]{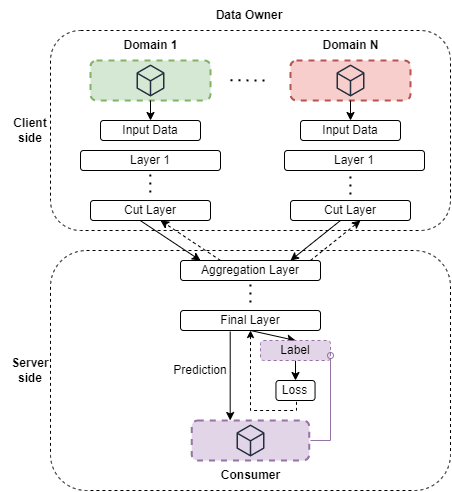}
    \caption{Distributed Domain Data with Label Sharing}
    \label{fig:sys1}
\end{figure}

\subsection{Distributed Domain Data without Label Sharing}

The second scenario arises when label data is considered sensitive, and its sharing with consumers is not permissible. This label data might either be safeguarded by an autonomous data owner or reside with one of the data domain teams. Under this architectural design, loss calculation is executed on the client side. Following this, the calculated loss is relayed back to the output layer situated at the final layer of the server model. Notably, the process of back-propagation of the gradient in this setup follows a U-shaped trajectory.

\begin{figure}[htbp]
    \centering
    \includegraphics[width=0.40\textwidth]{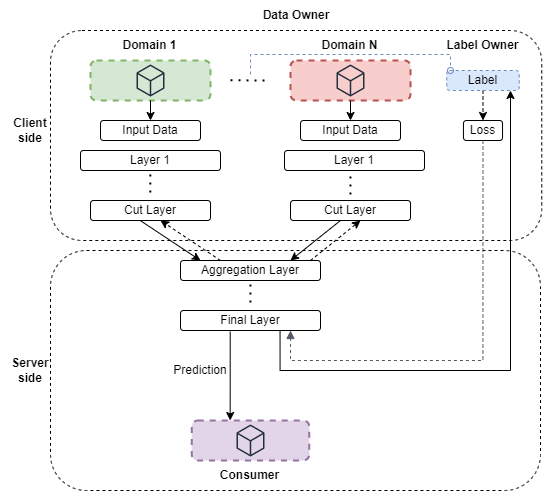}
    \caption{Distributed Domain Data without Label Sharing}
    \label{fig:sys2}
\end{figure}

\section{Use Cases}

To demonstrate the versatility of our proposed solution, we have constructed two prevalent business use cases necessitating the involvement of distributed domains.

\subsection{Recommendation System for Retail Industry}
\begin{figure}[htbp]
    \centering
    \includegraphics[width=0.45\textwidth]{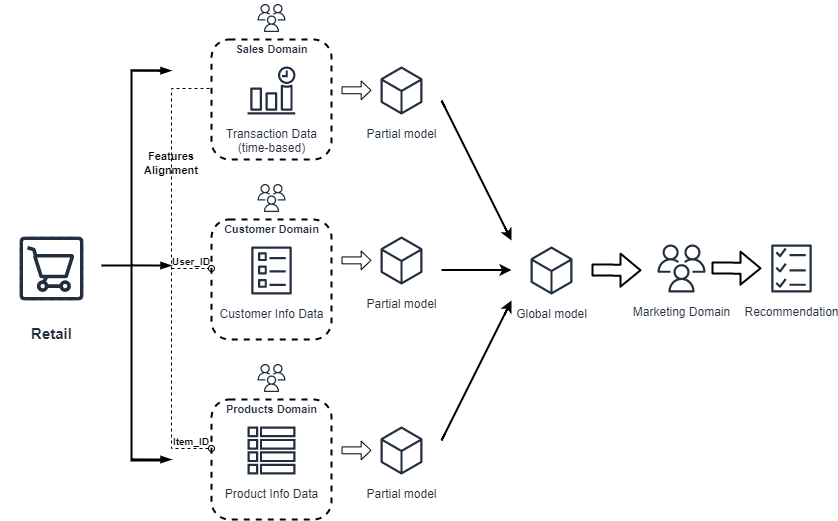}
    \caption{Recommendation System for Retail Industry}
    \label{fig:usecase_1}
\end{figure}

The first use case we implemented in our study is a personalized recommendation system in the retail industry, as is shown in Figure \ref{fig:usecase_1}. The H\&M Group\cite{hm_personalized_fashion} provided open-source dataset is used for this use case. It is inherently partitioned into three distinct data domains: transaction data (historical purchase records), article data (product information), and customer data (user metadata). This use case pertains to real-world retail businesses and their challenges associated with large-scale data management and analytics. 

The overarching task, as proposed by the federated governance team, is to construct a recommendation system. Each domain thus contributes by generating a partial model based on its proprietary data. These models then serve as upstream data products that are used by the marketing team to build a comprehensive recommendation model for subsequent analyses, such as optimizing marketing investment.

\subsection{Fraud Detection for Financial Institution}
Fraud detection presents a security-sensitive concern within financial organizations, often requiring a careful balance of data security and analytical accessibility across multiple domains. In our study, we utilize an anonymized credit card transaction dataset\cite{credit_card_fraud}, which is hypothetically partitioned into three distinct domains: finance, cardholder, and security. In this setup, the fraud prevention domain is tasked with detecting fraudulent activities, despite lacking direct access to raw data. 
Consequently, each data domain collaborates by contributing a partial model towards this effort. 

\begin{figure}[htbp]
    \centering
    \includegraphics[width=0.45\textwidth]{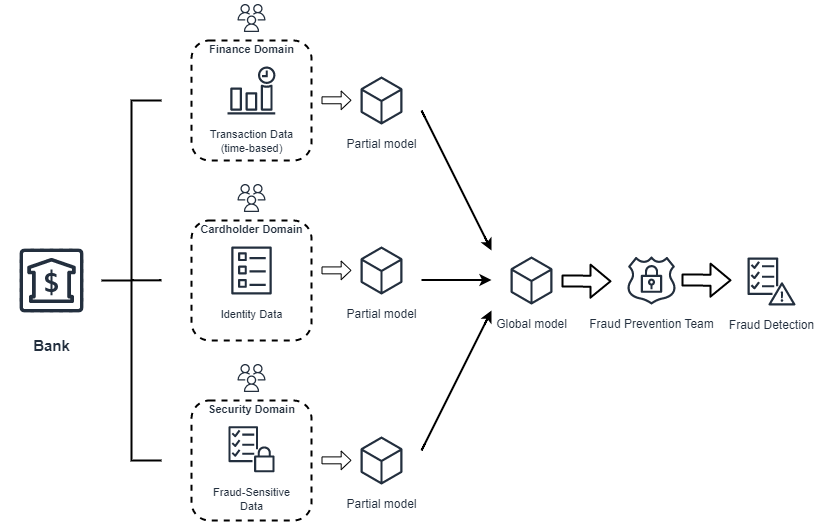}
    \caption{Fraud Detection for Financial Institution}
    \label{fig:usecase_2}
\end{figure}

This use case is closely aligned with organizations managing online transactions, necessitating an exceptionally high level of security and the adept handling of sensitive data. The federated governance team holds the responsibility of supervising the training process, in addition to formulating and issuing policies pertaining to encryption techniques, thereby ensuring the security of the intermediate models. The high-level architecture of this use case is plotted in Figure \ref{fig:usecase_2}.

\section{Results and Discussion}

 The following experiments are carried out on the above-mentioned two use cases: recommendation systems for fashion retailing and credit card fraud detection in banks. Each use case possesses a centralized model and a split learning model. The centralized model represents the scenario when data is stored in a data center and owned by the central data team. This is also the typical architecture for the current enterprise data platform. The split learning model creates the scenario when the data is distributed to decentralized data domains. In this case, the ownership of domain data belongs to the corresponding data team within the data mesh.\par

In the decentralized configuration, data will be loaded by a distributed DataLoader. To simulate the structure of the data mesh, features are split into different domains and sent out to each data owner. Domain teams provide a data product containing all autonomous technical components (e.g., code, metadata, operational API, and infrastructure) to serve other domain teams and global decision-making\cite{datameshblog2020,datameshblog2019, MACHADO2022263}. Here, specifically, is the partial model provided by each data domain team.

\subsection{Environment Setup}
All experiments are conducted on the Linux server, equipped with an Intel Xeon Gold $6230R$ Processor, $128GB$ RAM, and an NVIDIA RTX A5000 GPU with $24$ GB of memory. 

To address the needs of the split learning we have used PySyft as a base framework \cite{ziller2021pysyft}. PySyft is an open-source library created to facilitate privacy-preserving deep learning. 

All programs are written in Python $\text{3.7.12}$ using the PySyft library (PySyft $\text{0.2.9}$) and compatible PyTorch library (PyTorch $\text{1.4.0}$). Using JupyterLab for interactive deployment and visualization. The source code is available in a public GitHub repository \footnote{\url{https://github.com/Haoyuan-L/Fed_DataMesh}}.

\subsection{Datasets and required partitions}

Two public datasets are used in experiments. The first dataset, H\&M Personalized Fashion Recommendations, is provided by H\&M Group for product recommendation based on previous purchases \cite{hm_personalized_fashion}. 
In total, the dataset comprises $1.37$ million users, 106k products, and 31.8 million transaction records. 
Three sizes of datasets—small, medium, and large—are generated based on sampling ratios of $0.001$, $0.003$, and $0.01$, respectively. The distribution of three datasets is summarized in Table \ref{tab:hm_dataset}.\par

The NCF (Neural Collaborative Filtering) model learns implicit feedback from transaction data, where only the positive class (e.g., customer-product interactions) is observed\cite{he2017neural}. To address the inherent one-class problem, the negative sampling (NEG) technique is adopted\cite{mikolov2013distributed}. NEG balances the implicit dataset by generating a set of negative sampling from the unseen user-item matrix. This prevents model over-fitting on positive interaction and reduces the computation complexity\cite{mikolov2013distributed}. The NEG ratio is set to $5$ for small and medium size of data, and $2$ for large size of data, as recommended by \cite{mikolov2013distributed}.

\begin{table}
\centering
\small
\caption{Samples in H\&M Personalized Fashion Recommendations}
\label{tab:hm_dataset}
\begin{tabular}{
    @{}
    l
    c
    c
    c
    c
    @{}
}
\toprule
\textbf{Dataset} & {\textbf{Training}} & {\textbf{Testing}} & {\textbf{No. Users}} & {\textbf{No. Items}} \\
\midrule
H\&M-small  &   78,246 & 10,109 &  919 &  1,132 \\
H\&M-medium &  393,312 & 34,683 & 3,153 &  4,807 \\
H\&M-large  & 1,097,915 & 246,360 & 8,212 & 25,773 \\
\bottomrule
\end{tabular}
\end{table}

The second selected dataset is anonymous credit card transaction records provided by Worldline and the Machine Learning Group of ULB\cite{credit_card_fraud}.  Features $V1$ through $V28$ are derived from Principal Component Analysis (PCA), while transaction time and the amount have not undergone any transformation. A notable aspect of this dataset is its highly imbalanced nature in terms of fraud detection. As is shown in table \ref{imbalance}, the imbalance ratio between the negative and positive classes stands at a mere $0.17\%$. This imbalance tends to bias the model towards the majority class and hinders its ability to discern patterns within the minority class.\par

In our study, we adopt the Synthetic Minority Oversampling Technique (SMOTE)\cite{smote} to relieve the effect of bias in the data. We initially utilized SMOTE to augment the data points in the minority class. SMOTE generates synthetic samples in the feature space of the minority class based on the  $Euclidean$ distance. A large sample ratio may lead to the model overfitting the minority class; thus, we set the sample ratio as $1.5\%$. 
To further optimize the classifier's performance across both classes, we combine the random under-sampling with SMOTE, a practice endorsed by the original authors of the SMOTE algorithm\cite{smote}. The distribution of the resampled data is visualized in Table \ref{imbalance}.

\begin{table}[h]
\centering
\caption{Comparison of Class Imbalance Ratio}
\label{imbalance}
\begin{tabular}{@{}lS[table-format=6.0]S[table-format=4.0]S[table-format=1.2, table-auto-round, table-omit-exponent]@{}}
\toprule
\textbf{Dataset} & {\textbf{Negative Class}} & {\textbf{Positive Class}} & {\textbf{Imbalance Ratio (\%)}} \\
\midrule
Original & 199020 & 344 & 0.17 \\
Sampled & 149250 & 2985 & 2.0 \\
\bottomrule
\end{tabular}
\end{table}

\subsection{Evaluation Metrics}

In the use case of the recommendation system,  the accuracy metric is not appropriate to measure the quality of the ranking. Here, we select Hit Ratio at $K$ (HR@K),  and Normalized Discounted Cumulative Gain at $K$ ($NDCG@K$) to evaluate the performance of random recommendations in $k$ items. We use Recall at $K$ (Recall@K) to assess the ability of the model to find all the relevant cases within the test dataset. Hit rate measures the proportion of cases where the true item was among the top $K$ items in the ranked list. The definition is given as follows:
\begin{equation}
HR@K = \frac{1}{N} \sum_{i=1}^{N} rel_i
\end{equation}
$N$ is the total number of users, $rel$ is an indicator function. In the implicit data, $rel_i$ is $1$ if the actual interacted item is within the top $K$ predicted items for the $i^{th}$  user, and $0$ otherwise. In our experiments, the length of the recommendation list $K$ is set as $10$. \par

For the evaluation of our ranking model's performance, we employ the $NDCG@K$, a robust metric that weighs the position of relevant items within the ranked list. $NDCG$ is calculated by dividing the Discounted Cumulative Gain ($DCG$) of the presented ranked list by the $DCG$ of the ideally ranked list ($IDCG$). Notably, within the context of our research, each user has only one item with which they have actually interacted. This item, we propose, should ideally occupy the premier position in the ranking. As a result, the value of $IDCG_i$ in our specific context is $1$.

\begin{equation}
NDCG@K = \frac{1}{N} \sum_{i=1}^{N} \frac{DCG_i}{IDCG_i} = \frac{1}{N} \sum_{i=1}^{N} {DCG_i}
\end{equation}
where $DCG_i$ is given by:
\begin{equation}
DCG_i = \sum_{j=1}^{K} \frac{rel_{ij}}{log_2(j+1)}
\end{equation}
The range of NDCG is 0 to 1. A higher score signifies a better model. \par

In our research, we employ $Recall@K$ as another performance metric. This metric measures the fraction of actual interacted items correctly included in the top $K$ recommendations provided by the model. This metric proves particularly valuable in the context of implicit feedback datasets, where the model's objective is to infer users' interests or preferences, which have not been explicitly indicated. The definition is given as:

\begin{equation}
\text{Recall@K} = \frac{\text{\# of interacted items in top K predicted items}}{\text{\# of interacted items}}
\end{equation}

In the context of fraud detection, the focus is often on identifying the minority class - the fraudulent transactions - which typically represent a small fraction of the total transactions. Therefore, the accuracy is not an ideal metric as it can be misleading. We also choose precision, recall, and the F1 score to measure the quality of the model.

\subsection{Accuracy Analysis}

Table \ref{tab:performance_comparison} represents the performance of the recommendation system on three sizes of the H\&M dataset to test the stability of the split neural network on the scaling dataset. The experiment is conducted on three system configurations: Recommendation System on Centralized Neural Network (CRN), Recommendation System on Split Neural Network with label sharing (SRN1), and Recommendation System on Split Neural Network without label sharing (SRN2).


\begin{table*}
\centering
\small
\caption{Performance Comparison of Three Models on Three Datasets}
\label{tab:performance_comparison}
\small 
\begin{tabular*}{\textwidth}{
    l
    @{\extracolsep{\fill}} 
    *{9}{S[table-format=1.3]} 
}
\toprule
 & \multicolumn{3}{c}{\textbf{H\&M-small}} & \multicolumn{3}{c}{\textbf{H\&M-medium}} & \multicolumn{3}{c}{\textbf{H\&M-large}} \\
 \cmidrule(lr){2-4} \cmidrule(lr){5-7} \cmidrule(lr){8-10}
Metric & {CRN} & {SRN1} & {SRN2} & {CRN} & {SRN1} & {SRN2} & {CRN} & {SRN1} & {SRN2} \\
\midrule
NDCG@10 & 0.408 & \textbf{0.413} & 0.411 & \textbf{0.415} & 0.402 & 0.398 & 0.401 & \textbf{0.405} & 0.399 \\
HR@10 & 0.905 & 0.910 & \textbf{0.913} & \textbf{0.912} & 0.903 & 0.898 & \textbf{0.901} & 0.897 & 0.887 \\
Recall@10 & \textbf{0.900} & 0.868 & 0.890 & \textbf{0.959} & 0.943 & 0.925 & \textbf{0.982} & 0.956 & 0.960 \\
\bottomrule
\end{tabular*}
\end{table*}

In the recall aspect, CRN outperformed across all datasets, suggesting its effectiveness in not missing relevant recommendations. Yet, the performance of SRN1 or SRN2 is not drastically inferior. The potential benefits of decreased data movement in Split Learning models remain attractive alternatives, especially in larger, more complex architectures such as Data Mesh. In the context of implicit data, however, with relevance scores being binary (0 for unpurchased items and 1 for purchased), the prediction probability may not directly reflect the user's latent preferences. This difficulty in accurately predicting a user's preferences often results in a lower $NDCG$ score. This lower score does not necessarily imply poor model performance but highlights the inherent complexity of implicit feedback-based preference prediction. Neural Networks, as a parametric method, usually require abundant data points to learn the underlying pattern within the data. Therefore, we sampled three sizes of datasets from the original data to see the stability of model performance under the different scales of data. It turns out that our solutions maintain a consistent performance pattern even training on the small data. This result also aligns with the Data Mesh principle of decentralized, scalable data products.
\par

The results in Table \ref{tab:model_performance} explain the performance dynamics of the Centralized Fraud Neural Network (FraudNN) and the Split Fraud Neural Networks (with and without label sharing) in fraud detection. Despite a slight decrease in precision for the minority class, the Split Learning models' overall performance indicates a promising direction for machine learning in a Data Mesh framework, particularly for sensitive and critical applications such as fraud detection. \par

\begin{table*} 
    \centering
    \small
    \caption{Performance Metrics of Fraud Detection Models}
    \begin{subtable}{.33\linewidth}
        \centering
        \caption{Centralized FraudNN}
        \begin{tabular}{lccc}
            \toprule
            Metric & Class 0 & Class 1 & Macro Avg \\
            \midrule
            Precision & 1.00 & 0.82 & 0.91 \\
            Recall & 1.00 & 0.87 & 0.94 \\
            F1-score & 1.00 & 0.82 & 0.92 \\
            Support & 85295 & 148 & 85443 \\
            \bottomrule
        \end{tabular}
    \end{subtable}%
    \begin{subtable}{.33\linewidth}
        \centering
        \caption{Split FraudNN with Label Sharing}
        \begin{tabular}{lccc}
            \toprule
            Metric & Class 0 & Class 1 & Macro Avg \\
            \midrule
            Precision & 1.00 & 0.81 & 0.90 \\
            Recall & 1.00 & 0.86 & 0.93 \\
            F1-score & 1.00 & 0.82 & 0.92 \\
            Support & 85295  & 148 & 85443 \\
            \bottomrule
        \end{tabular}
    \end{subtable}%
    \begin{subtable}{.33\linewidth}
        \centering
        \caption{Split FraudNN without Label Sharing}
        \begin{tabular}{lccc}
            \toprule
            Metric & Class 0 & Class 1 & Macro Avg \\
            \midrule
            Precision & 1.00 & 0.83 & 0.91 \\
            Recall & 1.00 & 0.84 & 0.92 \\
            F1-score & 1.00 & 0.83 & 0.92 \\
            Support & 85295  & 148 & 85443 \\
            \bottomrule
        \end{tabular}
    \end{subtable}
    \label{tab:model_performance}
\end{table*}

Overall, the results illustrate the promising potential of Split Learning in the context of data mesh. It maintains competitive performance as data scales, which aligns with Data Mesh's principle of decentralized, scalable data products. Furthermore, the reduced data movement inherent in Split Learning models can offer significant advantages in Data Mesh.

\subsection{Diversity Analysis}

\begin{figure}[htbp]
    \centering
    \includegraphics[width=0.85\linewidth]{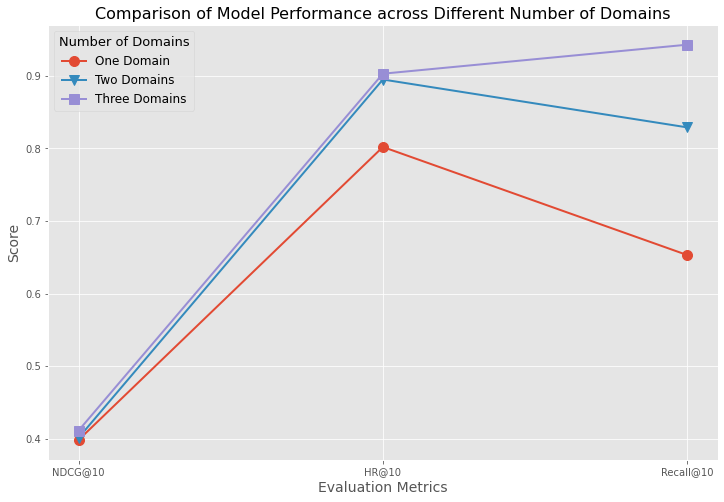}
    \caption{Model Performance across Multiple Domains in Recommendation System}
    \label{fig:diversity_exp1}
\end{figure}

\begin{figure}
    \centering
    \includegraphics[width=0.85\linewidth]{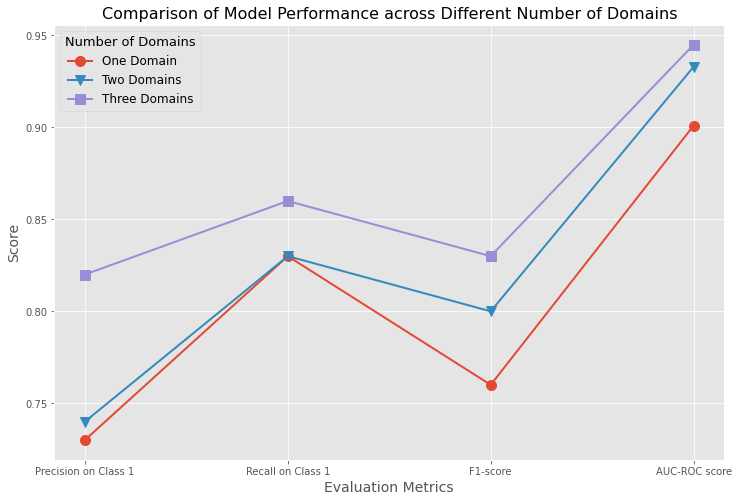}
    \caption{Model Performance across Multiple Domains in Fraud Detection}
    \label{fig:diversity_exp2}
\end{figure}

The second experiment builds upon the configuration of a Split Neural Network with label sharing, with a focus on scaling the number of data domains. The result are shown in Figure \ref{fig:diversity_exp1} and \ref{fig:diversity_exp2}. As we incrementally increased the number of data domains involved in the Split Neural Network model, we consistently observed improvements in both binary classification metrics (Precision, Recall, F1-score, and AUC-ROC score for Class 1) and recommendation system metrics (NDCG@10, HR@10, and Recall@10). The results infer that the autonomy and diverse characteristics of the domains within the structure of the data mesh, each contributing unique and valuable information, significantly boost the model's predictive acuity. \par

As we can see from the figures \ref{fig:diversity_exp1} and \ref{fig:diversity_exp2}, the results infer that the domains within the Data Mesh are able to boost the models' predictive performance by contributing unique and valuable information. This is because Data Mesh encourages multiple domain collaborations without revealing raw data. Moreover, the principle of distributed data ownership enables domain teams to fully harness the multifaceted data under the environment of Data Mesh. In conclusion, this experiment underscores the capability of Split Learning in leveraging the inherent diversity and decentralization of the Data Mesh architecture. Specifically, in data-sensitive scenarios like fraud detection, Split Learning empowers the domain-specific data teams to fully harness the multifaceted data within the Data Mesh.

\section{Conclusion}
In our work, we explore the integration of FL methodologies within the framework of Data Mesh. We examined different federated learning strategies, assessing their alignment with the architectural principles of Data Mesh. Additionally, we designed two configurations of split learning to address use cases involving both label sharing and label preservation. 
Our proposed methodologies empower the data domains within a data mesh to generate data products for consumers without necessitating the sharing of raw data. However, a potential risk lies in the exposure of intermediate data representations during the training phase, which could potentially lead to data leakage. Going forward, there should be a keen focus on safeguarding these intermediate data products. A variety of encryption techniques could be adopted to augment the security of the system.
This paper aims to serve as an applied groundwork, inspiring further scholarly pursuits to explore the incorporation of federated learning within the Data Mesh paradigm, thereby advancing the development of more secure and robust machine learning applications.

\begin{acks}
To the National Academic Infrastructure for Supercomputing in Sweden \cite{naiss} for cloud resources, eSSENCE strategic collaboration for support, and Assistant Professor Prashant Singh for technical discussions.
\end{acks}


\bibliographystyle{ACM-Reference-Format}

\begin{thebibliography}{30}


\ifx \showCODEN    \undefined \def \showCODEN     #1{\unskip}     \fi
\ifx \showDOI      \undefined \def \showDOI       #1{#1}\fi
\ifx \showISBNx    \undefined \def \showISBNx     #1{\unskip}     \fi
\ifx \showISBNxiii \undefined \def \showISBNxiii  #1{\unskip}     \fi
\ifx \showISSN     \undefined \def \showISSN      #1{\unskip}     \fi
\ifx \showLCCN     \undefined \def \showLCCN      #1{\unskip}     \fi
\ifx \shownote     \undefined \def \shownote      #1{#1}          \fi
\ifx \showarticletitle \undefined \def \showarticletitle #1{#1}   \fi
\ifx \showURL      \undefined \def \showURL       {\relax}        \fi
\providecommand\bibfield[2]{#2}
\providecommand\bibinfo[2]{#2}
\providecommand\natexlab[1]{#1}
\providecommand\showeprint[2][]{arXiv:#2}

\bibitem[int(2023)]%
        {intuitURL}
 \bibinfo{year}{(August 18, 2023)}\natexlab{}.
\newblock \bibinfo{title}{Intuit}.
\newblock \bibinfo{howpublished}{\url{https://www.intuit.com/}}.
\newblock


\bibitem[net(2023)]%
        {netflixURL}
 \bibinfo{year}{(August 18, 2023)}\natexlab{}.
\newblock \bibinfo{title}{Netflix}.
\newblock \bibinfo{howpublished}{\url{https://netflix.com}}.
\newblock


\bibitem[pay(2023)]%
        {paypalURL}
 \bibinfo{year}{(August 18, 2023)}\natexlab{}.
\newblock \bibinfo{title}{PayPal}.
\newblock \bibinfo{howpublished}{\url{https://www.paypal.com/}}.
\newblock


\bibitem[zal(2023)]%
        {zalandoURL}
 \bibinfo{year}{(August 18, 2023)}\natexlab{}.
\newblock \bibinfo{title}{zalando}.
\newblock \bibinfo{howpublished}{\url{https://zalando.com}}.
\newblock


\bibitem[Baker(2023)]%
        {intuitdatamesh}
\bibfield{author}{\bibinfo{person}{Tristan Baker}.} \bibinfo{year}{2022 (September 19, 2023)}\natexlab{}.
\newblock \bibinfo{title}{Intuit’s Data Mesh Strategy}.
\newblock \bibinfo{howpublished}{\url{https://medium.com/intuit-engineering/intuits-data-mesh-strategy-778e3edaa017}}.
\newblock


\bibitem[Beyer and Laney(2012)]%
        {beyer2012importance}
\bibfield{author}{\bibinfo{person}{Mark~A Beyer} {and} \bibinfo{person}{Douglas Laney}.} \bibinfo{year}{2012}\natexlab{}.
\newblock \showarticletitle{The importance of ‘big data’: a definition}.
\newblock \bibinfo{journal}{\emph{Stamford, CT: Gartner}} (\bibinfo{year}{2012}), \bibinfo{pages}{2014--2018}.
\newblock


\bibitem[Bode et~al\mbox{.}(2023)]%
        {bode2023data}
\bibfield{author}{\bibinfo{person}{Jan Bode}, \bibinfo{person}{Niklas K{\"u}hl}, \bibinfo{person}{Dominik Kreuzberger}, {and} \bibinfo{person}{Sebastian Hirschl}.} \bibinfo{year}{2023}\natexlab{}.
\newblock \showarticletitle{Data Mesh: Motivational Factors, Challenges, and Best Practices}.
\newblock \bibinfo{journal}{\emph{arXiv preprint arXiv:2302.01713}} (\bibinfo{year}{2023}).
\newblock


\bibitem[Butte and Butte(2022)]%
        {butte2022enterprise}
\bibfield{author}{\bibinfo{person}{Vijay~Kumar Butte} {and} \bibinfo{person}{Sujata Butte}.} \bibinfo{year}{2022}\natexlab{}.
\newblock \showarticletitle{Enterprise Data Strategy: A Decentralized Data Mesh Approach}. In \bibinfo{booktitle}{\emph{2022 International Conference on Data Analytics for Business and Industry (ICDABI)}}. IEEE, \bibinfo{pages}{62--66}.
\newblock


\bibitem[Ceballos et~al\mbox{.}(2020)]%
        {ceballos2020splitnn}
\bibfield{author}{\bibinfo{person}{Iker Ceballos}, \bibinfo{person}{Vivek Sharma}, \bibinfo{person}{Eduardo Mugica}, \bibinfo{person}{Abhishek Singh}, \bibinfo{person}{Alberto Roman}, \bibinfo{person}{Praneeth Vepakomma}, {and} \bibinfo{person}{Ramesh Raskar}.} \bibinfo{year}{2020}\natexlab{}.
\newblock \bibinfo{title}{SplitNN-driven Vertical Partitioning}.
\newblock
\newblock
\showeprint[arxiv]{2008.04137}~[cs.LG]


\bibitem[Chawla et~al\mbox{.}(2002)]%
        {smote}
\bibfield{author}{\bibinfo{person}{Nitesh~V Chawla}, \bibinfo{person}{Kevin~W Bowyer}, \bibinfo{person}{Lawrence~O Hall}, {and} \bibinfo{person}{W~Philip Kegelmeyer}.} \bibinfo{year}{2002}\natexlab{}.
\newblock \showarticletitle{SMOTE: synthetic minority over-sampling technique}.
\newblock \bibinfo{journal}{\emph{Journal of artificial intelligence research}}  \bibinfo{volume}{16} (\bibinfo{year}{2002}), \bibinfo{pages}{321--357}.
\newblock


\bibitem[Databricks(2023)]%
        {zalando}
\bibfield{author}{\bibinfo{person}{Databricks}.} \bibinfo{year}{2020 (August 18, 2023)}\natexlab{}.
\newblock \bibinfo{title}{Data Mesh in Practice: How Europe's Leading Online Platform for Fashion Goes Beyond the Data Lake}.
\newblock \bibinfo{howpublished}{\url{https://www.youtube.com/watch?v=eiUhV56uVUc}}.
\newblock


\bibitem[Dehghani(2023b)]%
        {datameshblog2019}
\bibfield{author}{\bibinfo{person}{Zhamak Dehghani}.} \bibinfo{year}{2019 (accessed June 5, 2023)}\natexlab{b}.
\newblock \bibinfo{title}{How to Move Beyond a Monolithic Data Lake to a Distributed Data Mesh}.
\newblock \bibinfo{howpublished}{\url{https://martinfowler.com/articles/data\-monolith-to-mesh.html}}.
\newblock


\bibitem[Dehghani(2023a)]%
        {datameshblog2020}
\bibfield{author}{\bibinfo{person}{Zhamak Dehghani}.} \bibinfo{year}{2020 (accessed March 3, 2023)}\natexlab{a}.
\newblock \bibinfo{title}{Data Mesh Principles and Logical Architecture}.
\newblock \bibinfo{howpublished}{\url{https://martinfowler.com/articles/data\-mesh\-principles.html}}.
\newblock


\bibitem[Goedegebuure et~al\mbox{.}(2023)]%
        {goedegebuure2023data}
\bibfield{author}{\bibinfo{person}{Abel Goedegebuure}, \bibinfo{person}{Indika Kumara}, \bibinfo{person}{Stefan Driessen}, \bibinfo{person}{Dario Di~Nucci}, \bibinfo{person}{Geert Monsieur}, \bibinfo{person}{Willem-jan van~den Heuvel}, {and} \bibinfo{person}{Damian~Andrew Tamburri}.} \bibinfo{year}{2023}\natexlab{}.
\newblock \showarticletitle{Data Mesh: a Systematic Gray Literature Review}.
\newblock \bibinfo{journal}{\emph{arXiv preprint arXiv:2304.01062}} (\bibinfo{year}{2023}).
\newblock


\bibitem[Gupta and Raskar(2018)]%
        {gupta2018distributed}
\bibfield{author}{\bibinfo{person}{Otkrist Gupta} {and} \bibinfo{person}{Ramesh Raskar}.} \bibinfo{year}{2018}\natexlab{}.
\newblock \showarticletitle{Distributed learning of deep neural network over multiple agents}.
\newblock \bibinfo{journal}{\emph{Journal of Network and Computer Applications}}  \bibinfo{volume}{116} (\bibinfo{year}{2018}), \bibinfo{pages}{1--8}.
\newblock


\bibitem[He et~al\mbox{.}(2017)]%
        {he2017neural}
\bibfield{author}{\bibinfo{person}{Xiangnan He}, \bibinfo{person}{Lizi Liao}, \bibinfo{person}{Hanwang Zhang}, \bibinfo{person}{Liqiang Nie}, \bibinfo{person}{Xia Hu}, {and} \bibinfo{person}{Tat-Seng Chua}.} \bibinfo{year}{2017}\natexlab{}.
\newblock \showarticletitle{Neural collaborative filtering}. In \bibinfo{booktitle}{\emph{Proceedings of the 26th international conference on world wide web}}. \bibinfo{pages}{173--182}.
\newblock


\bibitem[Kaggle(2023b)]%
        {credit_card_fraud}
\bibfield{author}{\bibinfo{person}{Kaggle}.} \bibinfo{year}{2018 (accessed March 3, 2023)}\natexlab{b}.
\newblock \bibinfo{title}{Machine Learning Group - ULB, "Credit Card Fraud Detection}.
\newblock \bibinfo{howpublished}{\url{https://www.kaggle.com/datasets/mlg-ulb/creditcardfraud}}.
\newblock


\bibitem[Kaggle(2023a)]%
        {hm_personalized_fashion}
\bibfield{author}{\bibinfo{person}{Kaggle}.} \bibinfo{year}{2022 (accessed March 12, 2023)}\natexlab{a}.
\newblock \bibinfo{title}{H\&M Personalized Fashion Recommendations}.
\newblock \bibinfo{howpublished}{\url{https://www.kaggle.com/competitions/h\-and\-m\-personalized\-fashion\-recommendations/overview}}.
\newblock


\bibitem[Khine and Wang(2018)]%
        {khine2018data}
\bibfield{author}{\bibinfo{person}{Pwint~Phyu Khine} {and} \bibinfo{person}{Zhao~Shun Wang}.} \bibinfo{year}{2018}\natexlab{}.
\newblock \showarticletitle{Data lake: a new ideology in big data era}. In \bibinfo{booktitle}{\emph{ITM web of conferences}}, Vol.~\bibinfo{volume}{17}. EDP Sciences, \bibinfo{pages}{03025}.
\newblock


\bibitem[Machado et~al\mbox{.}(2021)]%
        {machado2021data}
\bibfield{author}{\bibinfo{person}{In{\^e}s Machado}, \bibinfo{person}{Carlos Costa}, {and} \bibinfo{person}{Maribel~Yasmina Santos}.} \bibinfo{year}{2021}\natexlab{}.
\newblock \showarticletitle{Data-driven information systems: the data mesh paradigm shift}.
\newblock  (\bibinfo{year}{2021}).
\newblock


\bibitem[Machado et~al\mbox{.}(2022)]%
        {MACHADO2022263}
\bibfield{author}{\bibinfo{person}{Inês~Araújo Machado}, \bibinfo{person}{Carlos Costa}, {and} \bibinfo{person}{Maribel~Yasmina Santos}.} \bibinfo{year}{2022}\natexlab{}.
\newblock \showarticletitle{Data Mesh: Concepts and Principles of a Paradigm Shift in Data Architectures}.
\newblock \bibinfo{journal}{\emph{Procedia Computer Science}}  \bibinfo{volume}{196} (\bibinfo{year}{2022}), \bibinfo{pages}{263--271}.
\newblock
\showISSN{1877-0509}
\urldef\tempurl%
\url{https://doi.org/10.1016/j.procs.2021.12.013}
\showDOI{\tempurl}
\newblock
\shownote{International Conference on ENTERprise Information Systems / ProjMAN - International Conference on Project MANagement / HCist - International Conference on Health and Social Care Information Systems and Technologies 2021}.


\bibitem[Mikolov et~al\mbox{.}(2013)]%
        {mikolov2013distributed}
\bibfield{author}{\bibinfo{person}{Tomas Mikolov}, \bibinfo{person}{Ilya Sutskever}, \bibinfo{person}{Kai Chen}, \bibinfo{person}{Greg~S Corrado}, {and} \bibinfo{person}{Jeff Dean}.} \bibinfo{year}{2013}\natexlab{}.
\newblock \showarticletitle{Distributed representations of words and phrases and their compositionality}.
\newblock \bibinfo{journal}{\emph{Advances in neural information processing systems}}  \bibinfo{volume}{26} (\bibinfo{year}{2013}).
\newblock


\bibitem[NAISS(2023)]%
        {naiss}
\bibfield{author}{\bibinfo{person}{NAISS}.} \bibinfo{year}{2018 (accessed March 3, 2023)}\natexlab{}.
\newblock \bibinfo{title}{National Academic Infrastructure for Supercomputing}.
\newblock \bibinfo{howpublished}{\url{https://www.naiss.se/}}.
\newblock


\bibitem[Netflix(2023c)]%
        {NetflixVideo}
\bibfield{author}{\bibinfo{person}{Netflix}.} \bibinfo{year}{2020 (August 18, 2023)}\natexlab{c}.
\newblock \bibinfo{title}{Netflix Data Mesh: Composable Data Processing}.
\newblock \bibinfo{howpublished}{\url{https://www.youtube.com/watch?v=TO_IiN06jJ4}}.
\newblock


\bibitem[Netflix(2023b)]%
        {netflixblog1}
\bibfield{author}{\bibinfo{person}{Netflix}.} \bibinfo{year}{2021 (August 18, 2023)}\natexlab{b}.
\newblock \bibinfo{title}{Data Movement in Netflix Studio via Data Mesh}.
\newblock \bibinfo{howpublished}{\url{https://netflixtechblog.com/data-movement-in-netflix-studio-via-data-mesh-3fddcceb1059}}.
\newblock


\bibitem[Netflix(2023a)]%
        {netflixblog2}
\bibfield{author}{\bibinfo{person}{Netflix}.} \bibinfo{year}{2022 (August 18, 2023)}\natexlab{a}.
\newblock \bibinfo{title}{Data Mesh: A Data Movement and Processing Platform}.
\newblock \bibinfo{howpublished}{\url{https://netflixtechblog.com/data-mesh-a-data-movement-and-processing-platform-netflix-1288bcab2873}}.
\newblock


\bibitem[Perrin(2023)]%
        {paypaldatamesh}
\bibfield{author}{\bibinfo{person}{Jean-Georges Perrin}.} \bibinfo{year}{2022 (November 11, 2023)}\natexlab{}.
\newblock \bibinfo{title}{The Next Generation of Data Platforms is the Data Mesh}.
\newblock \bibinfo{howpublished}{\url{https://medium.com/paypal-tech/the-next-generation-of-data-platforms-is-the-data-mesh-b7df4b825522}}.
\newblock


\bibitem[Vepakomma et~al\mbox{.}(2018)]%
        {vepakomma2018split}
\bibfield{author}{\bibinfo{person}{Praneeth Vepakomma}, \bibinfo{person}{Otkrist Gupta}, \bibinfo{person}{Tristan Swedish}, {and} \bibinfo{person}{Ramesh Raskar}.} \bibinfo{year}{2018}\natexlab{}.
\newblock \showarticletitle{Split learning for health: Distributed deep learning without sharing raw patient data}.
\newblock \bibinfo{journal}{\emph{arXiv preprint arXiv:1812.00564}} (\bibinfo{year}{2018}).
\newblock


\bibitem[Yang et~al\mbox{.}(2019)]%
        {yang2019federated}
\bibfield{author}{\bibinfo{person}{Qiang Yang}, \bibinfo{person}{Yang Liu}, \bibinfo{person}{Tianjian Chen}, {and} \bibinfo{person}{Yongxin Tong}.} \bibinfo{year}{2019}\natexlab{}.
\newblock \showarticletitle{Federated machine learning: Concept and applications}.
\newblock \bibinfo{journal}{\emph{ACM Transactions on Intelligent Systems and Technology (TIST)}} \bibinfo{volume}{10}, \bibinfo{number}{2} (\bibinfo{year}{2019}), \bibinfo{pages}{1--19}.
\newblock


\bibitem[Ziller et~al\mbox{.}(2021)]%
        {ziller2021pysyft}
\bibfield{author}{\bibinfo{person}{Alexander Ziller}, \bibinfo{person}{Andrew Trask}, \bibinfo{person}{Antonio Lopardo}, \bibinfo{person}{Benjamin Szymkow}, \bibinfo{person}{Bobby Wagner}, \bibinfo{person}{Emma Bluemke}, \bibinfo{person}{Jean-Mickael Nounahon}, \bibinfo{person}{Jonathan Passerat-Palmbach}, \bibinfo{person}{Kritika Prakash}, \bibinfo{person}{Nick Rose}, {et~al\mbox{.}}} \bibinfo{year}{2021}\natexlab{}.
\newblock \showarticletitle{Pysyft: A library for easy federated learning}.
\newblock \bibinfo{journal}{\emph{Federated Learning Systems: Towards Next-Generation AI}} (\bibinfo{year}{2021}), \bibinfo{pages}{111--139}.
\newblock


\end{thebibliography}

\end{document}